%% file: latex/main.tex
\pdfoutput=1
\documentclass[11pt]{article}

\usepackage[final]{acl}

\usepackage{times}
\usepackage{latexsym}
\usepackage{import}
\usepackage{verbatim}
\usepackage{amsthm,amsmath,amssymb}
\usepackage{mathrsfs}
\usepackage{algorithmic}
\usepackage{algorithm}
\usepackage{tabularx}
\usepackage{booktabs}
\usepackage{makecell}
\usepackage[most]{tcolorbox}
\tcbset{
  colback=gray!5!white,    % 灰白背景
  colframe=black!70,       % 深灰边框
  boxrule=0.5pt,
  arc=2pt,
  left=4pt,
  right=4pt,
  top=4pt,
  bottom=4pt
}
\usepackage[normalem]{ulem}

\usepackage[T1]{fontenc}
\usepackage[utf8]{inputenc}

\usepackage{microtype}
\usepackage{multirow}
\usepackage{inconsolata}
\usepackage{ulem}
\usepackage{graphicx}

\title{RECALL: REpresentation-aligned Catastrophic-forgetting ALLeviation via Hierarchical Model Merging}

\begin{sloppypar}
\renewcommand{\thefootnote}{\fnsymbol{footnote}}
\author{
    \textbf{\small Bowen Wang\textsuperscript{1,}\footnotemark[4]},
    \textbf{\small Haiyuan Wan\textsuperscript{1,2}\footnotemark[4]},
    \textbf{\small Liwen Shi\textsuperscript{1}},
    \textbf{\small Chen Yang\textsuperscript{5}},
    \textbf{\small Peng He\textsuperscript{1,2}},
    \textbf{\small Yue Ma\textsuperscript{1,2}},
    \textbf{\small Haochen Han\textsuperscript{2}}\\
    \textbf{\small Wenhao Li\textsuperscript{4}},
    \textbf{\small Tiao Tan\textsuperscript{1}},
    \textbf{\small Yongjian Li\textsuperscript{6}},
    \textbf{\small Fangming Liu\textsuperscript{2, 3,}\footnotemark[2]},
    \textbf{\small Yifan Gong\textsuperscript{2,}\footnotemark[2]},
    \textbf{\small Sheng Zhang\textsuperscript{1,}\footnotemark[2]}\\
    \textsuperscript{1}{\small Shenzhen International Graduate School, Tsinghua University}\\
    \textsuperscript{2}{\small Peng Cheng Laboratory,}
    \textsuperscript{3}{\small Huazhong University of Science and Technology}\\
    \textsuperscript{4}{\small Xiamen University},
    \textsuperscript{5}{\small The Hong Kong University of Science and Technology, Guangzhou}\\
    \textsuperscript{6}{\small School of Biomedical Engineering, Tsinghua University}\\
\small{
    \textbf{Correspondence:}
    \href{mailto:email@domain}{\{wangbw23, wanhy24\}@mails.tsinghua.edu.cn, fangminghk@gmail.com}
}\\
\small{
    \href{mailto:email@domain}{gongyf@pcl.ac.cn, zhang\_sh@mail.tsinghua.edu.cn}
}
}
\end{sloppypar}
\begin{document}
\begin{sloppypar}
\maketitle
\renewcommand{\thefootnote}{\fnsymbol{footnote}}
\footnotetext[4]{Equal contribution}
\footnotetext[2]{Corresponding authors}
\begin{abstract}
\import{sections/}{abstract.tex}

\end{abstract}

\section{Introduction}
\import{sections/}{Introduction_V3.tex}

%\section{Preliminaries}
%\import{sections/}{Preliminaries.tex}

% \section{Representation Dynamics in Layered LLMs}
\section{Empirical Observations of Representation Dynamics in LLMs}

\import{sections/}{Observation_V2.tex}

\section{RECALL: REpresentation-aligned Catastrophic-forgetting ALLeviation}
\import{sections/}{Our_Method.tex}

\section{Experiments}
\import{sections/}{Experiments_results.tex}

\section{Related works}
\import{sections/}{Related_works_V2.tex}

\section{Conclusions}
\import{sections/}{Conclusions.tex}

\section{Limitations}
\import{sections/}{Limitations.tex}

\section*{Acknowledgments}
\import{sections/}{Acknowledgments.tex}

\bibliography{Reference}

\appendix

% \textbf{TODO}\\
% \section{Example Appendix}

\label{sec:appendix}
\import{sections/}{Appendix.tex}

% This is an appendix.

% \import{sections/}{Experiments_setup.tex}

\end{sloppypar}
\end{document}

%% file: sections/abstract.tex
 % In this paper, we propose a **Hierarchical Model Merging (HMM)** method to address the challenges of catastrophic forgetting and knowledge fusion in large language models (LLMs) during continual learning. Our approach leverages data representation analysis across model layers to compute inter-model similarities, enabling layer-wise parameter merging that preserves prior knowledge while integrating new task-specific capabilities. By clustering representative samples from the training data and calculating cosine similarities between model representations at each layer, we derive dynamic merging weights via softmax normalization. This hierarchical strategy ensures that lower layers retain domain-specific knowledge (e.g., syntax and semantics) while higher layers adapt to task-specific patterns. Experimental results on diverse datasets (e.g., SST-2, SQuAD, MedMCQA) demonstrate that HMM outperforms baseline methods in mitigating performance degradation on previous tasks and achieving robust multi-domain generalization. Our work provides an efficient, data-free framework for knowledge consolidation in LLMs, offering practical insights into balancing continual adaptation and memory retention.

\renewcommand{\thefootnote}{\roman{footnote}}
We unveil that internal representations in large language models (LLMs) serve as reliable proxies of learned knowledge and propose \textbf{RECALL}\footnote[1]{https://github.com/bw-wang19/RECALL}, a novel representation-aware model merging framework for continual learning without access to historical data. RECALL computes inter-model similarity from layer-wise hidden representations over clustered typical samples, and performs adaptive, hierarchical parameter fusion to align knowledge across models. This design enables the preservation of domain-general features in shallow layers while allowing task-specific adaptation in deeper layers. Unlike prior methods that require task labels or incur performance trade-offs, RECALL achieves seamless multi-domain knowledge fusion and strong resistance to catastrophic forgetting. Extensive experiments across five NLP tasks and multiple continual learning scenarios show that RECALL outperforms baselines in both knowledge retention and generalization, providing a scalable and data-free solution for evolving LLMs.

%% file: sections/Introduction_V3.tex
% Large language models (LLMs) have achieved impressive progress in recent years, delivering significant breakthroughs in various downstream tasks, including question answering, text generation, mathematical reasoning, and becoming central to advanced AI applications such as chatbots, AI business agents, and recommendation systems~\citep{devlin2018bert,brown2020language,touvron2023llama,raffel2020exploring}. Typically, LLM training involves multiple stages: pre-training (PT) on vast general-domain corpora through unsupervised learning, followed by supervised fine-tuning (SFT) on labeled data specific to particular tasks or domains to adapt the model's general knowledge to specialized applications~\citep{brown2020language, touvron2023llama, wei2021finetuned, ouyang2022training}.

% Despite significant performance improvements through fine-tuning, a critical limitation of LLMs compared to human intelligence is their vulnerability to \textit{catastrophic forgetting} (CF). CF arises from data distribution shifts during training, leading to disruptive parameter updates that compromise prior learning~\citep{1989Catastrophic, 2016Overcoming, 8107520}. Given the increasing deployment of LLMs in continual and multi-domain scenarios, addressing CF is essential for developing robust models that combine specialized expertise with broad generalization~\citep{brown2020language, wei2021finetuned, achiam2023gpt, Doimo2024TheRL}.

Large language models (LLMs) have achieved impressive advances across tasks like question answering, text generation, and mathematical reasoning, powering applications such as chatbots, AI business agents, and recommendation systems~\citep{devlin2018bert,brown2020language,touvron2023llama,raffel2020exploring}. They are typically trained through unsupervised pre-training on large corpora, followed by supervised fine-tuning (SFT) on task-specific or domain-specific data~\citep{brown2020language, touvron2023llama, wei2021finetuned, ouyang2022training}. However, LLMs remain susceptible to \textit{catastrophic forgetting} (CF), where distribution shifts during training lead to parameter updates that overwrite prior knowledge~\citep{1989Catastrophic, 2016Overcoming, 8107520}. As LLMs are increasingly applied in continual and multi-domain settings, mitigating CF is essential to maintain both specialization and generalization~\citep{brown2020language, wei2021finetuned, achiam2023gpt, Doimo2024TheRL}.

\begin{comment}
Previous approaches addressing CF generally fall into two categories, each with inherent limitations:

\begin{itemize}
\item \textbf{Data-based methods} preserve past knowledge by revisiting stored samples from previous tasks during training on new tasks~\citep{NIPS2017_f8752278, 2016iCaRL, 2019Adversarial, Isele_Cosgun_2018}. These methods, however, require access to historical data, which can be impractical or raise privacy issues in real-world scenarios.
\item \textbf{Model-based methods} constrain model updates or isolate task-specific knowledge via regularization~\citep{2021Continual, 2016Overcoming, 8107520, wang-etal-2023-orthogonal} or architecture adaptation~\citep{rusu2016progressive, fernando2017pathnet, tian2024hydralora}. While flexible, they typically explore limited optimization spaces and struggle to effectively retain performance across diverse tasks.
\end{itemize}
\end{comment}

As illustrated in Figure~\ref{fig:cf-taxonomy}, previous approaches addressing CF generally fall into two categories, each with distinct strengths and limitations:

\begin{figure*}[t]
  \centering
  \includegraphics[width=1\textwidth]{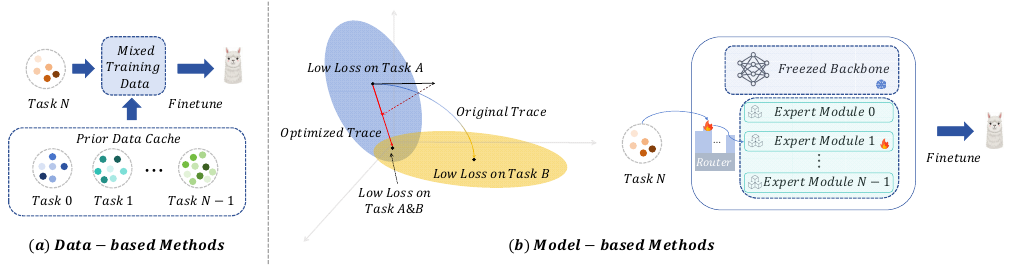}
  \caption{
A taxonomy of previous approaches to catastrophic forgetting. 
Data-based methods (a) rely on stored samples from previous tasks, which are replayed alongside new data during fine-tuning. 
Model-based methods (b) mitigate forgetting by either constraining parameter updates or isolating task-specific knowledge. 
In (b), the left side illustrates regularization-based methods that optimize model parameters within the intersection of low-loss regions for both old and new tasks (e.g., Task A and Task B), instead of strictly minimizing the loss on the new task. 
This encourages a more stable update trajectory that retains previously learned knowledge while adapting to new tasks.
}

  \label{fig:cf-taxonomy}
  \vspace{-1em}
\end{figure*}

% \begin{itemize}
% \item \textbf{Data-based methods} preserve past knowledge by revisiting stored samples from previous tasks during training on new tasks~\citep{NIPS2017_f8752278, 2016iCaRL, 2019Adversarial, Isele_Cosgun_2018}. These methods, however, require access to historical data, which can be impractical or raise privacy issues in real-world scenarios.
% \item \textbf{Model-based methods} constrain model updates or isolate task-specific knowledge via regularization~\citep{2021Continual, 2016Overcoming, 8107520, wang-etal-2023-orthogonal} or architecture adaptation~\citep{rusu2016progressive, fernando2017pathnet, tian2024hydralora}. While flexible, they typically explore limited optimization spaces and struggle to effectively retain performance across diverse tasks. Moreover, these approaches may rely on explicit task identifiers, which are often unavailable in real-world applications, and tend to increase model complexity during continual learning.
% \end{itemize}

% \begin{itemize}
% \vspace{-0.5em}

\noindent\textbf{1) Data-based methods} preserve past knowledge by revisiting stored samples from previous tasks during training on new tasks~\citep{NIPS2017_f8752278, 2016iCaRL, 2019Adversarial, Isele_Cosgun_2018}. These methods are effective in retaining task-specific information by directly exposing the model to prior data. However, they require access to historical samples, which may be impractical due to storage constraints or privacy concerns in real-world scenarios.
% \vspace{-0.5em}

\noindent\textbf{2) Model-based methods} constrain model updates or isolate task-specific knowledge via regularization~\citep{2021Continual, 2016Overcoming, 8107520, wang-etal-2023-orthogonal} or architecture adaptation~\citep{rusu2016progressive, fernando2017pathnet, tian2024hydralora}. These approaches enable continual learning without relying on past data, offering better scalability in privacy-sensitive settings. Nonetheless, they often operate within limited optimization spaces and struggle to preserve performance across diverse tasks. Additionally, they may depend on explicit task identifiers and increase model complexity over time.
% \vspace{-1.5em}
% \end{itemize}

% In addition to classical sequential continual learning scenarios, we also consider scenarios where multiple expert models are integrated into a single powerful general model~\citep{rusu2016progressive, fernando2017pathnet, tian2024hydralora, zhu2024llama}. In this work, we aim to leverage the interpretability of LLMs to provide a more intuitive approach. Prior research indicates a strong connection between CF and shifts in data representation within neural networks~\citep{9156964, 9857197, 2023Investigating, wang2024embedding, yu2025representation}. We further analyze representation drift occurring between transformer model layers, discovering that intermediate representations effectively capture essential task semantics. Based on this insight, we propose representing model capabilities using differences between these intermediate representations.

To overcome the limitations of existing continual learning approaches, we aim to combine the strengths of both data-based and model-based methods: retaining prior knowledge without relying on stored data, while enabling flexible model adaptation across tasks. %Yet, this integration is non-trivial. %The two paradigms present fundamentally different trade-offs: data-based methods rely on inaccessible training samples~\citep{NIPS2017_f8752278, Isele_Cosgun_2018}, while model-based approaches are constrained by limited optimization flexibility~\citep{rusu2016progressive, fernando2017pathnet, wang-etal-2023-orthogonal, tian2024hydralora}.

% However, (1)without access to historical data, it becomes difficult to assess what knowledge should be preserved; (2)and without explicit task boundaries, it is unclear how to guide model updates in a structured and generalizable manner. This raises a core challenge: \textbf{how can we identify and preserve useful task knowledge across models in a data-free and task-agnostic way?}
% Building on these challenges and inspired by recent advances in model merging techniques~\citep{xiao2023lm, wortsman2022model, jiang2023llm}, 
% observation->
% (1)By leveraging internal representations instead of raw data, we avoid the privacy and accessibility limitations of data-based approaches. (2)At the same time, our similarity-guided integration operates at a finer granularity than conventional model-based methods, substantially expanding the parameter search space and enabling more flexible and effective knowledge fusion.
% we propose a novel representation-aware model merging strategy that addresses both data availability and optimization flexibility issues. Specifically, our method computes inter-model similarities based on intermediate representations and uses them to guide adaptive, layer-wise parameter merging. 

However, without access to historical data, it becomes difficult to assess what knowledge should be preserved; and without explicit task boundaries, it is unclear how to guide model updates in a structured and generalizable manner. This raises a core challenge: \textbf{how can we identify and preserve useful task knowledge across models in a data-free and task-agnostic way?}

In addressing this question, we observe that internal representations, which reflect how models encode and process inputs, can serve as reliable proxies for their learned knowledge. These representations are inherently shaped by both model architecture and training objectives, making them well-suited for comparing and aligning knowledge across models without requiring access to raw data or task labels.
\newline
Motivated by this insight and recent advances in model merging~\citep{xiao2023lm, wortsman2022model, jiang2023llm}, we propose a novel representation-aware model merging strategy that addresses both data availability and optimization flexibility. Our method computes inter-model similarities based on intermediate representations and uses them to guide adaptive, layer-wise parameter merging. By avoiding raw data, we circumvent privacy and accessibility concerns, while our fine-grained integration expands the optimization space beyond traditional methods and enables more effective knowledge fusion.
\newline
\paragraph{Our main contributions are summarized as follows:}
\begin{itemize}
\item We propose a novel representation-aware model merging framework to address catastrophic forgetting, by leveraging intermediate representations to guide parameter fusion without relying on raw data or explicit task boundaries.
\vspace{-1em}
\item Our method generalizes to the merging of multiple expert models fine-tuned on different domains, enabling effective multi-domain capability fusion through weighted representation alignment.
\vspace{-1em}
\item We further demonstrate that the proposed framework can be applied to traditional continual learning benchmarks, including sequential fine-tuning scenarios, achieving strong performance without task-specific modifications.
\vspace{-2em}
\item Extensive experiments across multiple datasets and benchmarks validate the effectiveness and generality of our approach, showing consistent improvements in knowledge retention and transferability.
\end{itemize}

%% file: sections/Observation_V2.tex
Prior studies have shown that different layers of large language models encode distinct types of linguistic and semantic information~\citep{Tenney2019BERTRT, starace2023probing}. Building on this, we analyze hidden representations from transformer layers to examine how they evolve within a model and diverge across models fine-tuned on different tasks.

\subsection{Layer-wise Representation Shift}
\begin{figure}[t]
  \centering
  \includegraphics[width=\linewidth]{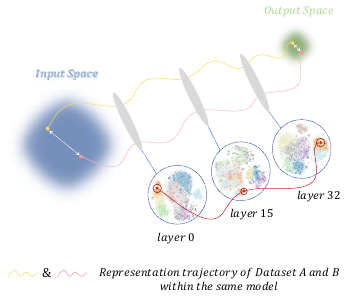}
  \caption{Illustration of representation transformation across layers within a single LLM. The input progresses through a sequence of transformations, and the corresponding hidden states (shown for layers 0, 15, and 32) exhibit distinct structural patterns in the representation space, highlighting the non-uniform nature of internal dynamics.}
  \label{fig:layer_tsne}
  \vspace{-1em}
\end{figure}

We first investigate how internal representations evolve across layers within a single model for a fixed input batch. Specifically, we compute the average RBF kernel similarity between adjacent layers’ hidden states. The similarity scores exhibit a non-monotonic pattern, with noticeable drops in both early and late layers. This indicates that the transformation of representations varies significantly across the network (see Appendix~\ref{appendix:sim-curve} for details).
In addition, as shown in Figure~\ref{fig:layer_tsne}, visualization through clustering and dimensionality reduction techniques shows that hidden states at different layers form distinct structural patterns in the representation space.

This layer-wise variation suggests that each layer contributes differently to the model’s behavior.
As a result, treating all layers uniformly during model merging—such as through naive parameter averaging—may overlook the unique functional roles of different layers and lead to suboptimal integration.

\subsection{Specialization-induced Model Divergence}

We next examine how internal representations diverge across models that share the same architecture and initialization but have been fine-tuned on different tasks. Using the same input batch, we extract hidden states from each model and compute the average layer-wise RBF kernel similarity between them.

\begin{figure}[t]
  \centering
  \includegraphics[width=\linewidth]{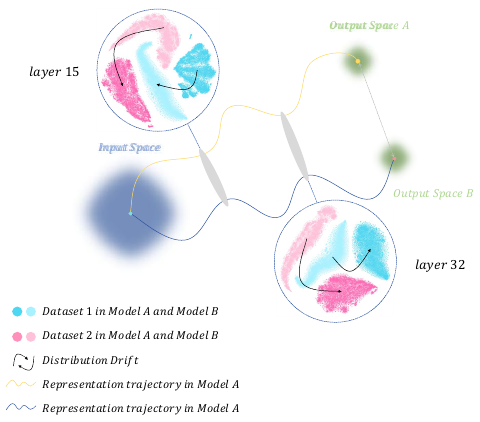}
  \caption{Visualization of representation drift between two models fine-tuned on different tasks (SST2 vs. RACE). Despite sharing the same input, their hidden states evolve along different trajectories and form distinct clustering patterns, especially in deeper layers.}
  \label{fig:inter_tsne}
  \vspace{-1em}
\end{figure}

We observe that lower-layer representations remain relatively consistent, while deeper layers diverge significantly across tasks—a trend highlighted by the model-wise similarity curves in Appendix~\ref{appendix:sim-curve}.

To further illustrate this phenomenon, Figure~\ref{fig:inter_tsne} visualizes the hidden states from two task-specific models. Despite processing the same inputs, their hidden states evolve along different trajectories and form distinct clustering structures, reinforcing the view that fine-tuning induces semantic specialization in deeper layers.

% We refer to this phenomenon as \textit{specialization-induced divergence}, where domain adaptation causes asymmetric representational shifts across models. This observation aligns with prior findings that upper layers in LLMs are more sensitive to task-specific information~\citep{tighidet2024probing, kotha2023understanding}.

These results suggest that naive parameter merging, especially in upper layers, may introduce semantic inconsistency or destructive interference if such representational misalignment is ignored.

%% file: sections/Our_Method.tex
\begin{figure*}[ht!]
  \centering
  \includegraphics[width=\linewidth]{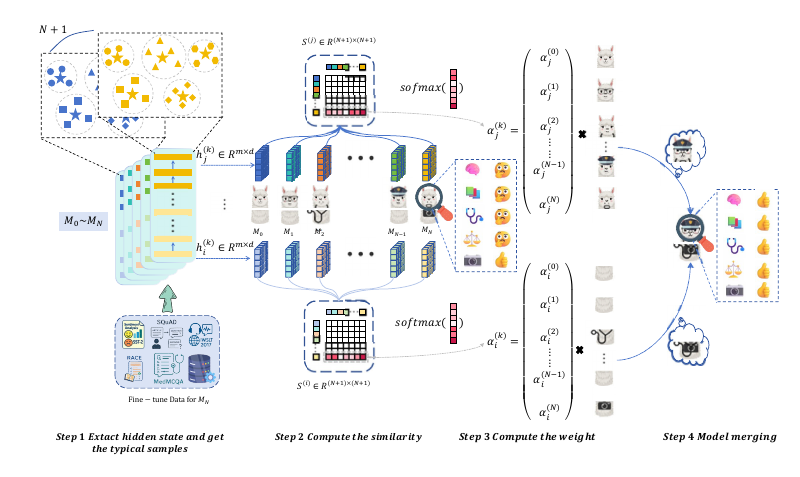}
  \vspace{-1em}
  \caption{Illustration of \textbf{RECALL}, our proposed representation-aware model merging framework. The pipeline consists of four stages: (1) extract hidden states from typical samples using the newly fine-tuned model $M_N$, (2) compute the pairwise representational similarities across all models (including $M_N$), (3) derive layer-wise adaptive weights based on similarity scores via softmax, and (4) perform hierarchical parameter merging guided by the computed weights. This process enables effective knowledge fusion across models while preserving task-specific features.}
  \label{fig:main_pipeline}
  \vspace{-1em}
\end{figure*}
\begin{comment}
    介绍从observation->method的过程
    ---
    在第三节中，我们通过实验观察，分析了数据表征在模型间、层间的特征，并发现模型的知识与数据表征的相似度密切相关。由此，我们在本节提出xxx，通过比较不同模型间数据表征的相似度，进行整体/分层式的模型融合，以强化模型在多个领域和任务的能力并抵抗灾难性遗忘。
\end{comment}
In Section 2, we analyze the characteristics of the data representation across models and layers through experimental observations, and illustrate that the knowledge of a model is closely related to its data representation. And previous works (\citealp{wortsman2022model}; \citealp{xiao2023lm}) have nicely illustrated that knowledge fusion and continual learning do not necessarily require a fine-tuning stage such as knowledge distillation. Model merging can also directly and effectively achieve the goal. 

Therefore, inspired by those observations, we propose RECALL in this section, which performs layer-wise model merging by comparing the similarities of data representations between different models, so as to achieve representation alignment. As illustrated in Figure~\ref{fig:main_pipeline}, RECALL effectively enhances LLM's abilities in multiple domains and tasks, and mitigates catastrophic forgetting.

As a prerequisite condition, we have the source model $M_0$ and multiple homologous expert models $M_1, M_2,\cdots,M_{N-1}$, which have the same architecture but different parameters with $M_0$. On the new task $T_N$, we obtain the new model $M_N$ by fine-tuning $M_0$ from the dataset $D_N$. Then, we select $m$ typical samples $D_{type}=\{d_1, d_2, \cdots, d_m\}\subset D_N$ through a clustering algorithm. For each $d_k\in D_{type}$, we extract its representations on models $M_{0}\sim M_N$, analyze the differences between $M_N$ and other models in semantic and syntactic knowledge through the similarities between  data representations. Finally, we perform hierarchical model merging for knowledge fusion.

\subsection{Data Representation}
\begin{comment}
    
\end{comment}
\textbf{Representation Extraction}. We extract the hidden states of layer $n$ of $M_p$, which is formulated as: $\mathbf{R}_n=(\mathbf{r}_1,\mathbf{r}_2,\cdots, \mathbf{r}_L)\in \mathbb{R}^{L\times E}$, where $L$ is the number of input tokens, $E$ is the dimension of embedding vectors, and $\mathbf{r}_i$ indicates the embedding of the $i$th token. Referring to the practice of most embedding models (\citealp{reimers-gurevych-2019-sentence}; \citealp{2023arXiv230907597X}), we average the hidden states by token to obtain the representation vector: $\mathbf{r}=\frac{1}{L}\sum_{i=1}^L\mathbf{r}_i\in \mathbb R^{E}$.

\textbf{Typical Dataset Selection}. Our approach does not place restrictions on the composition of datasets, which means that samples from multiple domains and tasks may be included in $D_N$. Therefore, we cluster all data representations of $D_N$, and select $m$ samples which are nearest to the $m$ cluster centers $C=\{c_1,c_2,\cdots, c_m\}$ to form the typical dataset $D_{type}=\{d_{t_1}, d_{t_2}, \cdots, d_{t_m}\}\subset D_{N}$. For $k\in[1,m]$: $d_{t_k}=\arg\min_{d_i\in D_N}\vert\vert d_i-c_k\vert\vert_2$, in which $c_k$ is the $k$th cluster center clustered by Kmeans.
In order to reduce the number of samples needed to perform forward inference for data representation analysis, we use $D_{type}$ as the representative of $D_N$ to analyze the knowledge difference of models.

\subsection{Similarity Calculation}

\begin{comment}
 for each sample $d_k$ in the typical dataset $D_{type}$ and each model $M_p$, its data representation at layer $i$ is $\mathbf{r}_i^{p,k}\in\mathbb R^{E}$.如前所述，我们利用不同模型的数据表征差异表示模型知识的差。具体来说，我们使用余弦相似度作为表征向量之间相似度指标，并使用m个样本表征向量的相似度的加权平均作为模型之间的整体相似度。在实验中，我们设定所有的样本都具有相同的权重，所以模型相似度就是所有样本的表征相似度的代数平均。
\end{comment}
 For each sample $d_k$ in the typical dataset $D_{type}$ and each model $M_p$, its data representation at layer $i$ is $\mathbf{r}_i^{p,k}\in\mathbb R^{E}$. As mentioned above, we measure the difference in knowledge between models by data representations. Specifically, as we select typical samples by Kmeans which is closely related to the norm distance, RBF kernel function is adopted to measure similarity between representation vectors, and we calculate the algebraic average of similarities of all $m$ samples in $D_{type}$ as the overall similarity. We also carefully discuss the differences using different similarity measures and do experiments to compare them; results and discussions are detailed in Appendix~\ref{appendix:sim-metric}.  

The similarity between $M_p$ and $M_q$ on layer $i$ is formulated as:
\begin{small}
    \begin{equation}
    \label{Similarity}
    S_i^{p,q}=\frac{1}{m}\sum_{k=1}^{m}\exp{(-\frac{\vert\vert\boldsymbol{r}_i^{p,k}-\boldsymbol{r}_i^{q,k}\vert\vert_2^2}{2\sigma^2})} ,
\end{equation}
\end{small}
in which $\sigma$ is a scaling factor.

\subsection{Hierarchical Merging}
% \begin{itemize}
    % \item[$\bullet$]\textbf{Paradigm Approach}
% \subsubsection{Paradigm Approach}
    \begin{comment}
    为了后面的叙述，我们在这里总结模型融合的范式方法。目前的模型融合本质上是对模型参数的线性插值，这意味着对于模型中的每个参数\theta, 有：\theta^*=\sum_{i=1}^Nw_i\thet_i=\boldsymbol{w}^T\boldsymbol{\theta}
    \end{comment}
    
    For later narration, we summarize here the paradigm approach to model merging. Current model merging is essentially a linear interpolation of the model parameters, which means for each parameter $\theta$ in the model and merging weight $w$, we have:
    \begin{small}
        \begin{equation}
        \label{paradigm}
    \theta^*=\sum_{i=1}^Nw_i\theta_i=\boldsymbol{w}^T\boldsymbol{\theta},
    \end{equation}
    \end{small}
    in which $\boldsymbol{\theta}$ is the vector concatenated by parameter $\theta$ of different models, and $\boldsymbol{w}$ is the vector of corresponding weights.
    
    Furthermore, Eq~\ref{paradigm} can be easily extended to the case that a group of parameters correspond to the same weights. We can compute linear interpolations of multiple parameters at once via the inner product operation like the following equation,
    \begin{small}
        \begin{equation}
            \label{paradigm2}
            \boldsymbol{\theta}
            =
            \begin{bmatrix}
                \theta_1\\
                \theta_2
            \end{bmatrix}
            =
            \begin{bmatrix}
                \boldsymbol{w}^T\boldsymbol{\theta}_1\\
                \boldsymbol{w}^T\boldsymbol{\theta}_2
            \end{bmatrix}
            =
            \begin{bmatrix}
                \boldsymbol{\theta}_1^T\\
                \boldsymbol{\theta}_2^T
            \end{bmatrix}
            \boldsymbol{w} .
        \end{equation}
    \end{small}
% \subsubsection{Hierarchical Merging}
    \begin{comment}
        在3.2节中，我们给出了相似度的度量方式，可以衡量模型之间某一层输出的相似程度。为了在模型间对齐数据表征，我们利用相似度计算合并权重。对第i层的相似度进行Softmax归一化，则模型M_q在第i层的权重为：
        
    \end{comment}
    
    In Section 3.2, we present the similarity metric to measure the similarities of data representation between models. To align their representations, we use Softmax to normalize representation similarities as merging weights. Then the weight of $M_q$ in layer $i$ is as follows:
    \begin{small}
    \begin{equation}
    \label{weights2}
    w_i^q=\frac{\exp{S_i^{n,q}}}{\sum\limits_{p=0}^{N} \exp{S_{i}^{n,p}}} .
    \end{equation}
    \end{small}
% \end{itemize}
Therefore, we provide representation-aligned merging method for one layer:
\begin{small}
    \begin{equation}
    \label{mergeing2}
    \boldsymbol{\theta}^*_i=\sum_{q=0}^Nw_i^q\boldsymbol{\theta}^q_i=
    \begin{bmatrix}
        \boldsymbol{\theta}^{1}_i, \
        \boldsymbol{\theta}^{2}_i, \ 
        \cdots,\,
        \boldsymbol{\theta}^{N}_i
    \end{bmatrix}\boldsymbol{w}_i
    =\boldsymbol{\Theta}_i^T\boldsymbol{w}_i, 
\end{equation}
\end{small}
where $\boldsymbol{\theta}_i$ denote the model's parameters of layer $i$.

According to Eq~\ref{paradigm2},~\ref{mergeing2}, we perform hierarchical model merging layer by layer:
\begin{small}
    \begin{equation}
    \label{mergeing1}
    \boldsymbol{\theta}^*
    =
    \begin{bmatrix}
        \boldsymbol{\theta}^{*}_1 \\
        \boldsymbol{\theta}^{*}_2 \\ 
        \vdots\\
        \boldsymbol{\theta}^{*}_L
    \end{bmatrix}
    =\begin{bmatrix}
        \boldsymbol{\Theta}_1^T\boldsymbol{w}_1 \\
        \boldsymbol{\Theta}_2^T\boldsymbol{w}_2 \\ 
        \vdots\\
        \boldsymbol{\Theta}_L^T\boldsymbol{w}_L
    \end{bmatrix}=diag(\boldsymbol{\Theta}^T\boldsymbol{w}),
\end{equation}
\end{small}
in which $\boldsymbol{\theta}^*$ is the parameter vector of the final merging model. $\boldsymbol{\Theta}=[\boldsymbol{\Theta}_1, \boldsymbol{\Theta}_2, \cdots, \boldsymbol{\Theta}_L]$ is the parameter matrix of $M_{0\sim N}$, and $\boldsymbol{w}^T=[\boldsymbol{w}_1^T,\boldsymbol{w}_2^T,\cdots,\boldsymbol{w}_N^T]$ is the corresponding weight matrix.

As mentioned above, our method enhances the abilities of LLM in multi-domains and resists catastrophic forgetting by performing independent weight calculation between layers  and hierarchical merging operations. The detailed procedure of RECALL is presented in Algorithm~\ref{alg:layerwise_merging} in the Appendix ~\ref{appendix:algorithm}, and meanwhile we provide an analysis of runtime, memory usage, and scalability.

%% file: sections/Experiments_results.tex
\begin{comment}
在本节中，我们将详细介绍我们的实验设置以及主要实验的结果，并对十余年结果进行总结与分析。这些实验有力地证明了我们方法的优越性。
\end{comment}
\renewcommand{\thefootnote}{\roman{footnote}}
In this section, we will provide a detailed introduction to our implementation and the results of experiments, which are mainly composed of three main parts: Experimental Setup, Different Merging Scenarios, and Sequential Fine-tuning Scenario. Furthermore, we summarize and analyze the results of these experiments, which strongly prove the superiority of our method.

\subsection{Experimental Setup} 
{\bf Datasets.} Considering a challenging experimental setup in knowledge fusion and continual learning, we selected 5 datasets as targets from multiple domains and tasks, including text classification, single-choice questions, and text generation, which are SST-2(\citealp{socher-etal-2013-recursive}), SQuAD2.0(\citealp{rajpurkar-etal-2016-squad, rajpurkar-etal-2018-know}), MedMCQA(\citealp{pmlr-v174-pal22a}), RACE(\citealp{lai-etal-2017-race}) and IWSLT2017(\citealp{cettolo-etal-2017-overview}). Since these datasets come from different tasks and have different formats, in order to adapt our method, we unify them into \textbf{QA} format by constructing prompts. Examples of the prompts are accessible in Appendix~\ref{appendix:instruction-sample}.

\textbf{Baseline}.(1) \textbf{SFT only}: directly fine-tunes the base model on a single downstream task without considering any cross-task interactions or parameter sharing. (2) \textbf{Avg.}(\citealp{wortsman2022model}): averaging their parameters without any alignment or adjustment. (3) \textbf{DARE}(\citealp{Yu2023LanguageMA}): flexible strategy to combine with other baselines(Average or Task Vector method) and random dropout parameters. (4) \textbf{LM-Cocktail}(\citealp{xiao2023lm}): merges models by comparing loss on validation set. (5) \textbf{Task Vector}(\citealp{ilharco2022editing}): computes the difference between the base model and each fine-tuned model to perform add, subtraction, or interpolation to construct new task behaviors. (6) \textbf{EWC}(\citealp{2016Overcoming}): introduces a regularization term based on the Fisher Information Matrix to prevent forgetting.

We selected the Llama-2-7B-chat\footnote[1]{https://huggingface.co/meta-llama/Llama-2-7b-chat-hf}(\citealp{touvron2023llama}) as the base model for fine-tuning and weight merging on 8 NVIDIA V100 GPUs, and LoRA(\citealp{hu2022lora}) is deployed for the fine-tuning pipeline. The implementation details of the fine-tuning and evaluation pipeline are provided in Appendix~\ref{appendix:pipeline}. All implementation details are supplied in Appendix~\ref{appendix:datasets-and-sft}.
% \ref{appendix:datasets-and-sft}.
% To evaluate the performance of RECALL, we conducted experiments on these datasets and compared the results with several baselines. 

In experiments of comparing with other baselines, our method always uses the same setting: we select 20 typical samples for each layer by the clustering algorithm, and those samples are concatenated to form the typical dataset. Same as Eq~\ref{Similarity}, we adopt the RBF kernel function as the similarity, of which the scale factor $\sigma$ is set to $1.0$. Then we segment and calculate weights for each layer of the model to merge them independently(taking Llama2-7b-chat as an example, the model will have 33 different groups of merging weights).

\subsection{Performance of RECALL in Different Merging Scenarios}
Firstly, we fine-tune the base model on the above 5 datasets to obtain five corresponding expert models. We then set up two different scenarios depending on the number of models used in the merging, which will be illustrated in the next two subsections.
% (1) \textbf{Single Fine-tuned Model Merging}, and (2) \textbf{Multiple Fine-tuned Models Merging}.

% \begin{itemize}
%     \item[$\bullet$] \textbf{Single Fine-tuned Model Merging}
\subsubsection{Single Fine-tuned Model Merging}
    In this study, we consider the case of merging using a single fine-tuned model and its base model. With access to the training datasets for both models, we conduct comprehensive experiments to evaluate the proposed approach across different datasets. Our experiments compare the performance of several baselines using different datasets, and the results are presented in Table~\ref{Tab:merge_test1}.
\import{tables/}{merging_test1.tex}

    We draw the following observations from Table~\ref{Tab:merge_test1}:
    Our method \textsc{RECALL} consistently outperforms all baselines across diverse settings, achieving the highest average performance (\textbf{45.00}) and the best generalization to unseen tasks (\textbf{38.92}, \textbf{+7.86\% }over the best baseline). It maintains top-tier results across all fine-tuning sources and excels in challenging domains such as \textsc{MedMCQA} and \textsc{IWSLT2017-en-fr}, demonstrating both robustness and transferability. These results underscore the effectiveness of leveraging representational similarity for model merging and motivate the extension to more complex multi-source integration scenarios.

\subsubsection{Multiple Fine-tuned Models Merging}
    To simulate a more complex knowledge fusion setting, we simultaneously merge five task-specific expert models. As shown in Table~\ref{Tab:merge_test2}, we consider two configurations: merging with and without the inclusion of the base model.
    \import{tables/}{merging_test2.tex}

    From Table~\ref{Tab:merge_test2}, we observe:
    \textsc{RECALL} achieves the best overall performance in both settings, with or without the base model, reaching averages of \textbf{56.93} and \textbf{62.83}, respectively. Notably, it outperforms all other methods even without relying on the base model, demonstrating a strong capability in fusing knowledge from multiple fine-tuned experts. These results highlight the advantage of representation-aware merging over both parameter averaging and task-vector-based baselines, and demonstrate that \textsc{RECALL} is not only effective for single-expert scenarios but also scalable to multi-expert merging, showing robust performance in both knowledge preservation and generalization without requiring access to training data.

    To more effectively demonstrate the effectiveness of RECALL across multiple models, we conducted supplementary experiments using Qwen2-7B-Instruct\footnote[2]{https://huggingface.co/Qwen/Qwen2-7B-Instruct}. As shown in Table~\ref{Tab:merge_test3}, RECALL also demonstrated a strong knowledge fusion ability and the ability to resist catastrophic forgetting in this test.
    \import{tables/}{merging_test3.tex}

\subsection{Sequential Fine-tuning Scenario}
To further assess the effectiveness of RECALL in realistic continual learning settings, we conduct sequential fine-tuning experiments across five tasks introduced in a fixed order. After training on each new task, the current model is merged with the previously accumulated one using different strategies. We compare RECALL against two baselines: standard LoRA-based fine-tuning (LoRA SFT) and Elastic Weight Consolidation (EWC)~\citep{2016Overcoming}.

\begin{figure}[ht!]
    \centering
    \includegraphics[width=\linewidth]{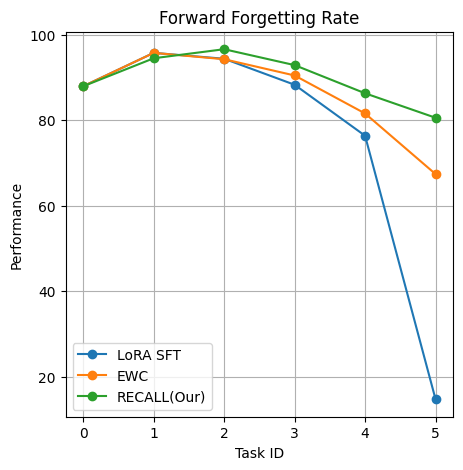}
    \caption{Performance curves on SST-2 during sequential fine-tuning with other two baselines.}
    \label{fig:forward_forgetting}
    \vspace{-1.5em}
\end{figure}

Figure~\ref{fig:forward_forgetting} illustrates the forward forgetting curves over the task sequence, where the y-axis indicates model performance on the current task immediately after learning it, and the x-axis denotes the task index.

 As illustrated in Figure~\ref{fig:forward_forgetting}, LoRA SFT suffers from a dramatic performance decline on the original SST-2 task as training progresses on new tasks, indicating a severe forward forgetting phenomenon. EWC alleviates this to some extent, but still shows a noticeable downward trend. In contrast, our proposed RECALL method maintains relatively stable performance throughout the sequential fine-tuning process, with only a moderate decline toward the final tasks. This suggests that RECALL is more effective at preserving prior task knowledge compared to the other two baselines.
 \newline
These results confirm that RECALL is well-suited for deployment in dynamic learning environments, offering resilience to forgetting while ensuring consistent learning progress. Detailed per-task results are available in Appendix~\ref{appendix:seq-ft}.

%% file: tables/merging_test1.tex
\begin{table*}[ht]
\resizebox{\linewidth}{!}{
\begin{tabular}{c|c|ccccc|cc}
\hline
\centering
\multirow{2}{*}{Fine-tuned on} & \multirow{2}{*}{Method} & \multicolumn{5}{c|}{Datasets}                          & \multirow{2}{*}{Avg from all tasks}    & \multirow{2}{*}{Avg from unseen tasks} \\ \cline{3-7}
                                                         &                         & SST-2  & SQuAD2.0 & IWSLT2017-en-fr & RACE   & MedMCQA &                             &                              \\ \hline
\multirow{5}{*}{SST-2}                                   & SFT only                & 95.76  & 31.68    & 13.28           & 44.71  & 32.32   & \uline{43.55}  & \uline{30.50} \\
                                                         & Avg                     & 94.95  & 5.21     & 12.32           & 32.75  & 34.28   & 35.90                      & 21.14                        \\
                                                         & DARE+Avg                & 95.07  & 8.75     & 11.68           & 50.44  & 35.14   & 40.22                      & 26.50                      \\
                                                         & LM-Cocktail             & 95.76  & 25.54    & 11.88           & 32.55  & 34.26   & 40.00                      & 26.06                      \\
                                                         & RECALL(Our)             & 94.50   & 30.72    & 12.08           & 47.44  & 34.90    & \textbf{43.93}             & \textbf{31.29}              \\ \hline
\multirow{5}{*}{SQuAD2.0}                                & SFT only                & 86.81  & 85.46    & 21.28           & 48.78  & 32.61   & \textbf{55.00}             & \uline{47.37}   \\
                                                         & Avg                     & 89.11  & 80.92    & 18.28           & 31.58  & 34.23   & 50.82                      & 43.3                         \\
                                                         & DARE+Avg                & 89.11  & 78.67    & 19.59           & 50.52  & 34.69   & 54.52                      & \textbf{48.48}             \\
                                                         & LM-Cocktail             & 79.36  & 84.46    & 18.38           & 42.24  & 32.66   & 51.42                       & 43.16                        \\
                                                         & RECALL(Our)             & 86.19  & 84.87    & 18.20            & 49.50   & 34.34   & \uline{54.62}  & 47.06                      \\ \hline
\multirow{5}{*}{IWSLT2017-en-fr}                         & SFT only                & 82.68  & 10.72    & 45.33           & 29.39  & 33.21   & 40.27                     & 39.00                           \\
                                                         & Avg                     & 89.91  & 4.35     & 42.01           & 32.85  & 35.45   & 40.91                      & 40.64                        \\
                                                         & DARE+Avg                & 89.33  & 10.45    & 41.63           & 44.6   & 35.84   & \uline{44.37}  & \uline{45.06}  \\
                                                         & LM-Cocktail             & 89.91  & 5.23     & 43.13           & 30.08  & 35.07   & 40.68                      & 40.07                      \\
                                                         & RECALL(Our)             & 89.56  & 10.55    & 43.09           & 48.73  & 34.43   & \textbf{45.27}             & \textbf{45.82}             \\ \hline
\multirow{5}{*}{RACE}                                    & SFT only                & 18.23      & 50.64    & 19.06           & 85.71  & 39.68   & 42.66                      & 31.90                       \\
                                                         & Avg                     & 47.36  & 14.80     & 22.58           & 73.47  & 34.66   & 38.57                      & 29.85                        \\
                                                         & DARE+Avg                & 29.13  & 50.05    & 19.55           & 78.68  & 34.97   & 42.48                      & 33.42                       \\
                                                         & LM-Cocktail             & 30.39  & 51.24    & 23.02           & 82.31  & 37.29   & \textbf{44.85}              & \textbf{35.49}              \\
                                                         & RECALL(Our)             & 34.93  & 40.27    & 23.12           & 79.31  & 36.96   & \uline{42.92} & \uline{33.82}   \\ \hline
\multirow{5}{*}{MedMCQA}                                 & SFT only                & 9.91      & 6.58     & 18.22           & 31.76  & 45.54   & 22.40                       & 16.62                        \\
                                                         & Avg                     & 0.11   & 5.97     & 15.34           & 23.17  & 43.25   & 17.57                      & 11.15                      \\
                                                         & DARE+Avg                & 24.36  & 11.99    & 14.04           & 57.46  & 42.86   & \uline{30.14} & \uline{26.96} \\
                                                         & LM-Cocktail             & 17.58  & 12.61    & 14.07           & 24.89  & 44.18   & 22.67                      & 17.29                      \\
                                                         & RECALL(Our)             & 70.32  & 18.58    & 13.86           & 43.77  & 44.82   & \textbf{38.27}              & \textbf{36.63}             \\ \hline
\multirow{5}{*}{All Average}                             & SFT only                & 58.68  & 37.02   & 23.43          & 48.07  & 36.67  & 40.77                     & 33.08                      \\
                                                         & Avg                     & 64.29 & 22.25    & 22.11          & 38.76 & 36.37  & 36.76                     & 29.22                      \\
                                                         & DARE+Avg                & 65.40   & 31.98   & 21.30          & 56.34  & 36.7    & \uline{42.34}                      & \uline{36.08}                      \\
                                                         & LM-Cocktail             & 62.60   & 35.82   & 22.10          & 42.40 & 36.69  & 39.92                     & 32.41                      \\
                                                         & RECALL(Our)             & 75.10   & 37.00   & 22.07           & 53.75  & 37.09   & \textbf{45.00}($+6.28\%$)                     & \textbf{38.92}($+7.86\%$)                      \\ \hline
\end{tabular}
}
\caption{Performance of merging the base model(Llama-2-7B-chat) and the model fine-tuned on one specific dataset. We compared our method with 4 baselines and marked the best two results in \textbf{bold} and \uline{underlined} fonts. The average performance on 5 datasets and 4 datasets(except the fine-tuning dataset) is also labeled in the last two columns.}
\label{Tab:merge_test1}
\vspace{-1em}
\end{table*}

%% file: tables/merging_test2.tex
% \usepackage{multirow}
\begin{table*}[ht]
\resizebox{\linewidth}{!}{
\begin{tabular}{cc|ccccc|c}
\hline
% \centering
\multicolumn{1}{l|}{\multirow{2}{*}{}}                   & \multirow{2}{*}{Method} & \multicolumn{5}{c|}{Datasets}                        & \multirow{2}{*}{Average} \\ \cline{3-7}
\multicolumn{1}{l|}{}                                    &                         & SST-2 & SQuAD2.0 & IWSLT2017-en-fr & RACE  & MedMCQA &                          \\ \hline
\multicolumn{2}{c|}{Llama2-7B-chat(base model)}                                    & 87.96 & 0.94     & 9.64            & 50.14 & 35.91   & 36.918                   \\ \hline
\multicolumn{1}{c|}{\multirow{4}{*}{With base model}}    & Avg.                    & 86.47 & 54.85    & 27.25           & 58.65 & 35.84   & 52.612                   \\
\multicolumn{1}{c|}{}                                    & DARE+Avg.               & 86.35 & 63.9     & 34.24           & 61.63 & 36.82   & \uline{56.588}             \\
\multicolumn{1}{c|}{}                                    & LM-Cocktail             & 51.38 & 66.74    & 29.31           & 68.89 & 36.07   & 50.478                   \\
\multicolumn{1}{c|}{}                                    & RECALL(Our)             & 85.44 & 78.4     & 28.26           & 57.9  & 34.66   & \textbf{56.932}          \\ \hline
\multicolumn{1}{c|}{\multirow{5}{*}{Without base model}} & Avg.                    & 91.28 & 67.85    & 35.87           & 66.94 & 37.2    & 59.828                   \\
\multicolumn{1}{c|}{}                                    & DARE+Avg.               & 89.6  & 68.01    & 36.85           & 69.08 & 40.96   & \uline{60.9}               \\
\multicolumn{1}{c|}{}                                    & Task Vector             & 11.82 & 29       & 9.98            & 49.64 & 7.36    & 21.56                    \\
\multicolumn{1}{c|}{}                                    & DARE+Task Vector        & 16.86 & 29.34    & 11.11           & 50.34 & 9.25    & 23.38                    \\
\multicolumn{1}{c|}{}                                    & RECALL(Our)             & 89.11 & 77.66    & 33.12           & 74.39 & 39.86   & \textbf{62.828}          \\ \hline
\end{tabular}
}
\caption{Performance of merging multiple models. \textbf{With base model}: Merging the five fine-tuned models and the base model(Llama-2-7B-chat). \textbf{Without base model}: Merging the five fine-tuned models. We compared our method with several baselines and marked the best two results in \textbf{bold} and \uline{underlined} fonts. }
\label{Tab:merge_test2}
\end{table*}

%% file: tables/merging_test3.tex
% \usepackage{multirow}

\begin{table*}[ht]
\resizebox{\linewidth}{!}{
\begin{tabular}{c|ccccc|c}
\hline
% \centering
\multirow{2}{*}{Fine-tuned on} & \multicolumn{5}{c|}{Datasets}                        & \multirow{2}{*}{Average} \\ \cline{2-6}
                               & SST-2 & SQuAD2.0 & IWSLT2017-en-fr & RACE  & MedMCQA &                          \\ \hline
Without SFT                    & 93    & 37.02    & 41.85           & 87.66 & 52.45   & 62.396                   \\ \hline
SST-2                          & 96.79 & 35       & 41.43           & 76.92 & 46.78   & 59.384                   \\ \hline
SQuAD2.0                       & 92.55 & 98.51    & 42.33           & 50.63 & 24.38   & 61.68                    \\ \hline
IWSLT2017-en-fr                & 92.32 & 31.49    & 52.62           & 86.34 & 50.9    & 62.734                   \\ \hline
RACE                           & 28.9  & 25.89    & 35.28           & 99.15 & 54.35   & 48.714                   \\ \hline
MedMCQA                        & 0.57  & 13.43    & 35.5            & 88.71 & 97.7    & 47.182                   \\ \hline
RECALL\_6merges                    & 94.15 & 81.45    & 45.03           & 91.93 & 59.14   & \textbf{74.34}       \\ \hline   
\end{tabular}
}
\caption{\textbf{Supplementary Experiments}: Performance of Qwen2-7B-Instruct and models fine-tuned on one specific dataset, compared with the model merged with the above six models using RECALL. The best result is marked in \textbf{bold} font.}
\label{Tab:merge_test3}
\vspace{-1em}
\end{table*}

%% file: sections/Related_works_V2.tex
% Large-scale transformer-based language models (LLMs) are increasingly deployed in multi-domain and continual learning scenarios. However, fine-tuning LLMs sequentially often results in \textit{catastrophic forgetting} (CF), where performance on previously learned tasks degrades as new knowledge is acquired. 
Catastrophic forgetting (CF) is particularly severe in realistic deployment settings, where training data from previous tasks may be inaccessible due to privacy concerns, and task boundaries or identifiers are typically unavailable. To address CF, existing continual learning (CL) approaches can be broadly categorized into two classes: \textbf{data-based methods} and \textbf{model-based methods}. Data-based methods leverage stored or generated samples from earlier tasks~\citep{NIPS2017_f8752278, 2016iCaRL}, while model-based methods impose constraints on parameter updates or isolate task-specific modules~\citep{2016Overcoming, fernando2017pathnet}. Some recent work adapts these paradigms to LLMs using parameter-efficient tuning modules~\citep{wei2025online, tian2024hydralora}.

\subsection{Model Merging}

Model merging has emerged as an alternative to traditional CL methods, enabling knowledge integration without access to historical training data. Most methods perform parameter-level fusion, typically via uniform averaging, without accounting for layer-wise functional differences.

Task Arithmetic~\citep{ilharco2022editing} and ModelSoup~\citep{wortsman2022model} showed that simple weight averaging can yield multi-task models. Fisher Merging~\citep{NEURIPS2022_70c26937} incorporates importance weights based on Fisher information to preserve task-relevant parameters. RegMean~\citep{jin2023dataless} formulates merging as a regression problem over model outputs, aligning them via low-rank projection.

Other works attempt to mitigate interference through more selective merging. TIES-Merging~\citep{yadav2023tiesmerging} trims parameter deltas and aligns signs, while DARE~\citep{Yu2023LanguageMA} sparsifies task-specific shifts to preserve key differences. LM-Cocktail~\citep{xiao2023lm} and LLM-Blender~\citep{jiang2023llm} perform weighted merging or output blending using learned domain signals or generation-based rankers.

\subsection{Probing Representations}

Probing techniques analyze how LLMs internally organize linguistic and task knowledge. Prior work has shown that lower layers tend to encode syntactic information, while upper layers capture semantics and abstract features~\citep{Tenney2019BERTRT, starace2023probing}.

\citet{starace2023probing} demonstrate that linguistic features are unevenly distributed across layers and can shift during adaptation. \citet{tighidet2024probing} find that past knowledge may remain latent but inaccessible, while \citet{kotha2023understanding} show that representation-level forgetting is limited, with performance loss arising from usage changes rather than loss of internal content.

These findings highlight the importance of analyzing internal representations when studying model behavior under adaptation and support representation-driven approaches to knowledge retention and integration.

%% file: sections/Conclusions.tex
\begin{comment}
在这项工作中，我们首先进行了探索性实验，从数据表征出发，探讨了数据表征在层与层之间和模型与模型之间产生偏移的现象，并将这种现象与模型的知识差异和灾难性遗忘联系起来。
借由这种发现，我们提出了一种通过对齐模型不同层的表征，来实现知识融合和抵抗灾难性遗忘的方法，也就是RECALL。RECALL不需要过往的数据，只需要利用模型表征的相似性进行分层式的模型聚合，就可以有效地实现目标。我们在多种场景下验证了该方法的有效性，并通过消融实验和其他测试对方法的细节进行了更深入的分析。有关我们对limitation的讨论 are available in section 8.
\end{comment}
In this work, we first conduct exploratory experiments to explore the phenomenon that data representations drift between layers and models, and relate this phenomenon to knowledge differences and catastrophic forgetting of models.
Based on these findings, we propose a method to achieve knowledge fusion and resist catastrophic forgetting by aligning the representations of different layers of the model, called RECALL. RECALL does not require past data and only requires hierarchical model aggregation by exploiting the similarity of model representations to achieve the goal effectively. We verify the effectiveness of the method in multiple scenarios, and analyze the details of the method in more depth through ablation experiments and other tests.

%% file: sections/Limitations.tex
While RECALL provides an effective and data-free solution to continual learning in large language models, several limitations remain. First, our method assumes access to multiple fine-tuned models on related tasks, which may not always be available in real-world deployment scenarios. Second, the current implementation relies on clustering and similarity computations over a small set of representative samples; while efficient, the selection quality of these typical samples can influence the final merging outcome. Moreover, RECALL is tailored to models with identical architectures and aligned tokenizers—extending to heterogeneous model families or multilingual settings poses additional challenges. Finally, although we empirically validate RECALL across diverse NLP tasks, further investigation is needed on scaling to dozens of tasks or integrating with training-time regularization techniques for tighter lifelong learning integration.

%% file: sections/Acknowledgments.tex
This work was supported in part by the Major Key Project of PCL under Grant PCL2024A06 and PCL2025AS10, and in part by the Shenzhen Science and Technology Program under Grant RCJC20231211085918010.

We would like to thank our colleagues and collaborators for their valuable feedback and insightful discussions throughout the course of this work. We are also grateful to the open-source community for providing access to pretrained language models and toolkits that significantly accelerated our research.

%% file: sections/Appendix.tex
\section{Datasets and Fine-Tuning Settings}
\label{appendix:datasets-and-sft}

We fine-tune the LLaMA-2-7B model on 5 different datasets from diverse domains and tasks, including sentiment classification, question answering, medical QA, reading comprehension, and machine translation. Detailed statistics and supervised fine-tuning (SFT) hyperparameters are presented below.

\subsection{Dataset Statistics and Prompt Format}
\begin{table}[ht]
\centering
\small
\begin{tabularx}{\linewidth}{Xccc}
\toprule
\textbf{Dataset} & \textbf{\# train} & \textbf{\# test} & \textbf{Metric} \\
\midrule
SST-2 & 60,000 & 872 & Accuracy \\
SQuAD2.0 & 130,000 & 11,873 & Exact Match \\
MedMCQA & 100,000 & 4,183 & Accuracy \\
RACE & 80,000 & 4,934 & Accuracy \\
IWSLT2017-en-fr & 100,000 & 8,597 & Exact Match \\
\bottomrule
\end{tabularx}
\caption{Statistics for the datasets used to fine-tune LLaMA-2-7B.}
\label{tab:fine-tune-datasets}
\end{table}

% \paragraph{Prompt Format.}
% Each instance is wrapped using the following unified prompt template:
% \begin{tcolorbox}[width=\linewidth]
% \small\ttfamily
% [INST] You are a helpful assistant. Given the following input, respond accordingly. Input: \{X\} [/INST]
% \end{tcolorbox}

\paragraph{Dataset Descriptions.}
\begin{itemize}
\item[$\bullet$] \textbf{SST-2}~\citep{socher-etal-2013-recursive}: Binary sentiment classification dataset with movie reviews labeled as positive or negative.
\item[$\bullet$] \textbf{SQuAD2.0}~\citep{rajpurkar-etal-2016-squad, rajpurkar-etal-2018-know}: Reading comprehension dataset with both answerable and unanswerable questions.
\item[$\bullet$] \textbf{MedMCQA}~\citep{pmlr-v174-pal22a}: Multiple-choice QA dataset from Indian medical entrance exams.
\item[$\bullet$] \textbf{RACE}~\citep{lai-etal-2017-race}: Reading comprehension dataset from English exams for Chinese middle and high school students.
\item[$\bullet$] \textbf{IWSLT2017-en-fr}~\citep{cettolo-etal-2017-overview}: English-to-French translation dataset from TED talks.
\end{itemize}

\subsection{Fine-Tuning Hyperparameters}

We fine-tune five task-specific models based on LLaMA-2-7B using LoRA~\citep{hu2022lora} on 8 NVIDIA V100 GPUs. Each model is trained with distinct hyperparameters tailored to its dataset. The LoRA config is reported as $r/\alpha/$dropout ((see Table~\ref{tab:sft-settings} for details)).

\begin{table*}[t]
\centering
\small
\setlength{\tabcolsep}{6pt}
\begin{tabular}{lcccccc}
\toprule
\textbf{Dataset} & \textbf{LoRA ($r/\alpha/$dropout)} & \textbf{Max Len} & \textbf{LR} & \textbf{Batch} & \textbf{Epochs} & \textbf{Deepspeed} \\
\midrule
SST-2           & 8 / 32 / 0.1 & 2048 & 5e-5 & 64 & 3 & ZeRO-3 \\
SQuAD2.0        & 8 / 32 / 0.1 & 2048 & 5e-5 & 32 & 4 & ZeRO-3 \\
MedMCQA         & 8 / 32 / 0.1 & 2048 & 5e-5 & 64 & 3 & ZeRO-3 \\
RACE            & 8 / 32 / 0.1 & 2048 & 5e-5 & 128 & 5 & ZeRO-3 \\
IWSLT2017-en-fr   & 8 / 32 / 0.1 & 2048 & 5e-5 & 64 & 5 & ZeRO-3 \\
\bottomrule
\end{tabular}
\caption{SFT hyperparameters for each dataset.}
\label{tab:sft-settings}
\end{table*}

\vspace{1em}
\noindent
The LLaMA-2-7B-chat~\citep{touvron2023llama} is used as the base model. We intentionally chose diverse datasets to simulate a challenging setup for continual learning and knowledge fusion.

\section{Experimental Framework}
\label{appendix:pipeline}

We adopt \texttt{llama-factory}~\citep{zheng2024llamafactory} for instruction tuning. It supports various parameter-efficient fine-tuning methods such as LoRA~\citep{hu2022lora} and QLoRA~\citep{dettmers2023qlora}, enabling flexible configuration and easy adaptation to various data formats.

Model performance is evaluated using \texttt{OpenCompass}~\citep{2023opencompass}, which integrates a broad range of standardized benchmarks to ensure consistency and reproducibility. For efficient inference, we deploy models with \texttt{vLLM}~\citep{kwon2023efficient}, providing high throughput and low latency.

% \section{Representation Clustering Visualization}
% \label{appendix:rep-cluster}

% To better understand how internal representations evolve across different layers and models, we visualize the hidden state representations using clustering-based dimensionality reduction. Specifically, we extract representations from \textbf{Layer 1}, \textbf{Layer 15}, and \textbf{Layer 31} of two fine-tuned models and project them into 2D space using t-SNE.

% Each plot corresponds to a specific model–layer pair, with color indicating the sample category (e.g., sentiment class or answerability). This allows us to observe how well the learned representations separate semantically different inputs and how this separation evolves across layers and between models.

% % \begin{figure}[h]
% %     \centering
% %     \includegraphics[width=0.48\linewidth]{figs/modelA_layer1.pdf}
% %     \includegraphics[width=0.48\linewidth]{figs/modelB_layer1.pdf} \\
% %     \includegraphics[width=0.48\linewidth]{figs/modelA_layer15.pdf}
% %     \includegraphics[width=0.48\linewidth]{figs/modelB_layer15.pdf} \\
% %     \includegraphics[width=0.48\linewidth]{figs/modelA_layer31.pdf}
% %     \includegraphics[width=0.48\linewidth]{figs/modelB_layer31.pdf}
% %     \caption{t-SNE visualization of hidden representations from Layer 1, 15, and 31 of two different models. Each color denotes a different sample category.}
% %     \label{fig:rep-clustering}
% % \end{figure}

\section{Instruction and Clustering Sample Details}
\label{appendix:instruction-sample}

To illustrate the data used in our experiments, we present two sets of representative samples. Table~\ref{tab:instruction-sample} shows instruction samples used during supervised fine-tuning (SFT).

\begin{table*}[ht]
\small
\centering
\begin{tabular}{p{1.5cm}p{10.5cm}}
\toprule
\textbf{Task} & \textbf{Example} \\
\midrule
\textbf{SST-2} & \textbf{Instruction:} Statement: the characters in swimfan seem motivated by nothing short of dull, brain-deadening hangover. What's sentiment should the above sentence be? OPTIONS:- negative.- positive. Answer: \newline
\textbf{Output:} negative \\
\addlinespace[0.7em]
\textbf{SQuAD2.0} & \textbf{Instruction:} Unpopulated boards are usually bare-board tested for ... the appropriate contact points and only on these. According to the above passage, answer the following question. If it is impossible to answer according to the passage, answer `impossible to answer`: Question:What\'s an absent connection that needs to be linked up on an unpopulated board called?\newline
\textbf{Output:} An open \\
\addlinespace[0.7em]
\textbf{MedMCQA} & \textbf{Instruction:} Question: Which of the following metabolic reactions require vitamin B12 but not folate? Options: A: Conversion of malonic acid to succinic acid B: Conversion of homocysteine to methionine C: Conversion of serine to glycine D: Thymidylate synthesis Choose an correct answer from A/B/C/D.Answer: \newline
\textbf{Output:} A \\
\addlinespace[1.0em]
\textbf{RACE} & \textbf{Instruction:} Read the article, and answer the question by replying A, B, C or D. Article: Tired of all the pushing in supermarkets? Angry at wasting ... claim it to be. Q:The author agrees with the fact that ... \newline
\textbf{Output:} D \\
\addlinespace[1.0em]

\bottomrule
\end{tabular}
\caption{Instruction samples used for supervised fine-tuning.}
\label{tab:instruction-sample}
\end{table*}

\vspace{1em}

% \begin{table*}[h]
% \small
% \centering
% \begin{tabular}{p{1.5cm}p{10.5cm}}
% \toprule
% \textbf{Task} & \textbf{Concatenated Input for Clustering} \\
% \midrule
% \textbf{SST-2} & \textbf{Input:} Determine the sentiment (positive/negative) of the following review: ``A touching and beautifully filmed story that will resonate with audiences of all ages.'' \newline
% \textbf{Output:} positive \newline
% \textbf{Concatenated:} Determine the sentiment (positive/negative) of the following review: “...” \texttt{<sep>} positive \\
% \addlinespace[0.7em]
% \textbf{SQuAD2.0} & \textbf{Input:} Passage: Marie Curie won two Nobel Prizes... \newline
% \textbf{Question:} How many Nobel Prizes did Marie Curie win? \newline
% \textbf{Output:} Two \newline
% \textbf{Concatenated:} Passage: ... Question: ... \texttt{<sep>} Two \\
% \addlinespace[0.7em]
% \textbf{MedMCQA} & \textbf{Input:} The organ responsible for detoxification in the human body is... \newline
% \textbf{Options:} (A)...(D) \newline
% \textbf{Output:} (C) Liver \newline
% \textbf{Concatenated:} The organ responsible for... (A)...(D) \texttt{<sep>} (C) Liver \\
% \bottomrule
% \end{tabular}
% \caption{Concatenated input-output samples used for hidden representation clustering.}
% \label{tab:clustering-sample}
% \end{table*}

\section{Supplementary Similarity Curves}
\label{appendix:sim-curve}

\begin{figure}[ht!]
    \centering
    \includegraphics[width=1\linewidth]{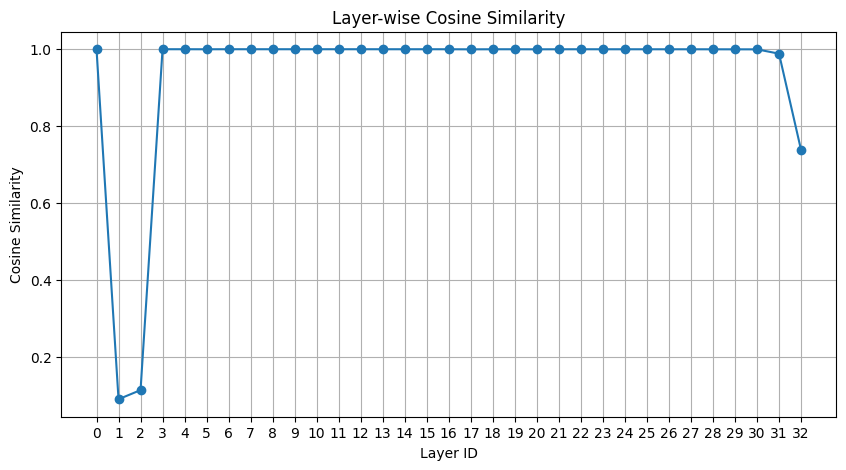}
    \caption{Cosine similarity between adjacent hidden layers within a single LLM. The similarity drops in both early and late layers, suggesting non-uniform transformation of representations across the network.}
    \label{fig:layer_wise_cos}
\end{figure}

\begin{figure}[ht!]
    \centering
    \includegraphics[width=1\linewidth]{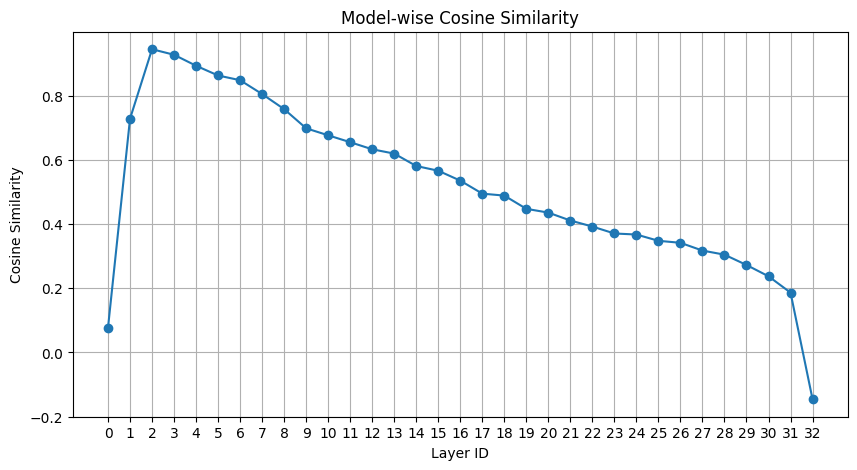}
    \caption{Cosine similarity between representations at the same layer across two LLMs fine-tuned on different tasks. Similarity remains high in early layers but decreases in deeper layers, indicating increasing task-specific divergence.}
    \label{fig:model_wise_cos}
\end{figure}

% To further validate the generality of our method, we present extended similarity curve results across additional models and datasets. These curves illustrate the layer-wise representational alignment between different models and serve to quantify how knowledge is preserved or shifted throughout the network.

% Specifically, we compute the cosine similarity between hidden representations extracted at each transformer layer. For each dataset, we visualize similarity trends across models fine-tuned on related or unrelated domains, providing insights into inter-model representational dynamics.

% \begin{figure*}[t]
%     \centering
%     \includegraphics[width=0.32\linewidth]{figs/sim_curve_modelA_vs_modelB_sst2.pdf}
%     \includegraphics[width=0.32\linewidth]{figs/sim_curve_modelA_vs_modelB_medmcqa.pdf}
%     \includegraphics[width=0.32\linewidth]{figs/sim_curve_modelA_vs_modelB_race.pdf} \\
%     \includegraphics[width=0.32\linewidth]{figs/sim_curve_modelB_vs_modelC_squad.pdf}
%     \includegraphics[width=0.32\linewidth]{figs/sim_curve_modelB_vs_modelC_iwslt.pdf}
%     \includegraphics[width=0.32\linewidth]{figs/sim_curve_modelB_vs_modelC_hc3.pdf}
%     \caption{Layer-wise cosine similarity between hidden representations from different fine-tuned models on six datasets. Top row: comparisons involving Model A and B. Bottom row: comparisons involving Model B and C.}
%     \label{fig:sim-curve-supp}
% \end{figure*}

As shown in Figure~\ref{fig:layer_wise_cos},~\ref{fig:model_wise_cos}, representational similarity varies across layers and tasks. Models trained on similar tasks (e.g., SST-2 and RACE) show higher alignment in middle and upper layers, while those from different domains (e.g., MedMCQA vs. IWSLT) diverge significantly, especially in deeper layers. These patterns are consistent with our main findings and further support the use of representation-aware merging strategies.

\section{Sequential Fine-Tuning Results}
\label{appendix:seq-ft}

To provide a strong baseline for comparison, we conduct sequential fine-tuning (SeqFT) experiments, where a single model is trained on multiple datasets in a fixed order without revisiting previous ones. This setting simulates a continual learning scenario and serves to quantify the extent of catastrophic forgetting.

We sequentially fine-tune the LLaMA-2-7B model across five diverse tasks, including sentiment classification, question answering, medical QA, reading comprehension, and machine translation. All models are trained under the same LoRA configuration for consistency. After completing each step in the sequence, we evaluate the model on all previously seen datasets to track performance drop.

\input{tables/sequence_test}

As shown in Table~\ref{Tab:sequence_test}, performance on earlier tasks gradually deteriorates as the model is updated on subsequent ones. The trend clearly reflects catastrophic forgetting and reinforces the need for continual learning strategies such as our proposed representation-aware model merging, which avoids overwriting previous knowledge by aligning and preserving internal representations.

\section{Comparison of Similarity Metrics}
\label{appendix:sim-metric}

To determine the most effective similarity metric for guiding our representation-aware model merging, we conduct a comparative study across five widely-used similarity measures. These metrics are used to compute the alignment between hidden representations of models, which in turn inform the layer-wise merging weights.

The five similarity metrics evaluated are:
\begin{itemize}
    \item[$\bullet$] \textbf{Cosine similarity}: $x,y$ are vectors.
    \begin{small}
    \begin{equation}
        Sim=\frac{x^Ty}{||x||_2\times||y||_2}
    \end{equation}
    \end{small}
    \item[$\bullet$] \textbf{Euclidean distance (converted to similarity)}: $x,y$ are vectors.
    \begin{small}
    \begin{equation}
        Sim=\frac{||x-y||_2}{\max\limits_{\mathcal{X,Y}}||\mathcal{X-Y}||_2}
    \end{equation}
    \end{small}
    \item[$\bullet$] \textbf{Centered Kernel Alignment (CKA)}~\citep{pmlr-v97-kornblith19a}: $X,Y$ are two distributions.
    \begin{small}
    \begin{equation}
\text{CKA}(X, Y) = \frac{\| X^\top Y \|_F^2}{\| X^\top X \|_F \cdot \| Y^\top Y \|_F}
\end{equation}
\end{small}
    \item[$\bullet$] \textbf{Maximum Mean Discrepancy (MMD)}: $X,Y$ are two distributions.
    \begin{small}
        \begin{equation}
        \begin{aligned}
        \mathrm{MMD}^2(X, Y) = \frac{1}{n^2} \sum_{i=1}^n \sum_{j=1}^n k(x_i, x_j) \\
        + \frac{1}{m^2} \sum_{i=1}^m \sum_{j=1}^m k(y_i, y_j) \\
        - \frac{2}{nm} \sum_{i=1}^n \sum_{j=1}^m k(x_i, y_j)
        \end{aligned}
        \end{equation}
    \end{small}
    \item[$\bullet$] \textbf{RBF kernel}: See Eq~\ref{Similarity}.
\end{itemize}

For each metric, we compute layer-wise alignment scores between expert models, normalize the weights, and perform hierarchical model merging using the same fusion strategy. The final merged models are evaluated on multiple tasks to assess performance consistency.

\begin{table*}[ht]
\centering
\small
\begin{tabular}{lcccccc}
\toprule
\textbf{Metric} & SST-2 & SQuAD2.0 & IWSLT2017-en-fr & RACE & MedMCQA & Avg. \\
\midrule
Cosine & 83.83 & 67.99 & 33.24 & 65.2 & 37.03 & 57.458\\
Euclidean & 88.65 & 26.72 & \textbf{43.64} & 38.93 & 34.71 & 46.53\\
CKA & 83.94 & 68.04 & 33.25 &65.16 & 36.91 & 57.46 \\
MMD & 65.83 & 28.93 & 41.26 & 50.87 & 36.58  & 44.694\\
RBF & \textbf{89.11} & \textbf{77.66} & 33.12 & \textbf{74.39} & \textbf{39.86}  & \textbf{62.828}\\
\bottomrule
\end{tabular}
\caption{Performance of merged multiple models(without base model) using different similarity metrics. RBF Kernel similarity consistently achieves the best average performance across tasks.}
\label{tab:sim-metric}
\end{table*}

As shown in Table~\ref{tab:sim-metric}, RBF Kernel yields the highest performance across all evaluation datasets. While CKA and dot product also perform competitively, metrics like Euclidean distance and MMD are less stable. These results support our choice of RBF Kernel as the default alignment metric in our model merging framework.

% \appendix
\section{RECALL Algorithm Details}
\label{appendix:algorithm}

\begin{algorithm}[!ht]
    \caption{\textbf{RECALL}}
    \label{alg:layerwise_merging}
    \begin{algorithmic}[1]
        \REQUIRE Task dataset $D_N$, source model $M_0$, fine-tuned models $M_1, M_2, \dots, M_{N-1}$ with parameters $\boldsymbol{\theta}^q \, (q \in [0, N-1])$
        \ENSURE Merged model parameters $\boldsymbol{\theta}^*$

        \STATE $M_N \leftarrow$ Fine-tune $M_0$ on $D_N$
        \STATE $D_{type}\leftarrow Kmeans(\mathbf{R}_N)$, which are representations extracted from $D_{N}$ using $M_N$ 
        
        \FOR{each expert model $M_j, \ j\in[1,N-1]$}    
            \STATE $\mathbf{R}_j \leftarrow$ Extract representations from $D_{type}$ using $M_j$
        \ENDFOR
        
        \FOR{each layer $i \in [1, L]$}
            \STATE Compute similarity $S_i^{p,q}$ between models $M_p$ and $M_q$ for layer $i$
            \[
                S_i^{p,q}=\frac{1}{||D_{type}||}\sum_{D_{type}}\mathbf{RBF}(R_p^i, R_q^i)
            \]
            \STATE Normalize similarities to obtain merging weights: 
            \[
                w_i^q = \frac{\exp(S_i^{N,q})}{\sum_{q=0}^{N} \exp(S_i^{N,q})}
            \]
            \STATE Merge model parameters at layer $i$: 
            \[
                \boldsymbol{\theta}_i^* = \sum_{q=0}^N w_i^q \boldsymbol{\theta}_i^q
            \]
        \ENDFOR

        \RETURN $\boldsymbol{\theta}^*$
    \end{algorithmic}
\end{algorithm}

\textbf{The Analysis of Runtime, Memory Usage and Scalability}

Compared with other model merging methods, our method mainly adds the following steps: (1) Feature extraction; (2) Extracting typical samples; (3) Extracting the representation of typical samples in all models; (4) Similarity calculation; (5) Hierarchical merging.

Mathematical notation convention: $t$: number of iterations; $k$: number of clusters; $n$: number of samples; $E$: dimension of features/hidden layers; $l$: number of model layers; $b$: mini-batch size; $s$: number of GPUs; $m$: number of typical samples; $N$: number of models to merge.
\begin{itemize}
    \item[$\bullet$] \textbf{Feature extraction}: Feature extraction forwards all samples and saves all hidden layer states. In order to save memory and speed up, we use distributed inference and pass features back to rank0 on each batch, where they are offloaded to CPU memory to reduce GPU memory usage. So the GPU memory complexity is $O(bEl)$, and the CPU memory and subsequent storage complexity is $O(nEl)$. And the time complexity is: $O(\alpha l\frac{n}{bs}+\beta nEl\frac{s-1}{s}+\gamma nEl)$, where the first term is the time required to forward samples across GPUs in parallel, The second term is the time it takes for sub nodes to send a batch of features to rank 0, the third term is the time it takes to offload all the features from GPU to CPU, and $\alpha,\beta,\gamma$are scaling constants. \textbf{If this distributed offloading strategy isn't adopted, the time complexity and space complexity will greatly increase}: Space complexity: $O(nEl)$; Time complexity: $O(\alpha nl)$. 
    \item[$\bullet$] \textbf{Extracting typical samples}: We perform Kmeans clustering for the features of each layer in step (1), and the space complexity and time complexity of the one-pass Kmeans algorithm are $O(E(n+k))$and $O(tknE)$, respectively. The overall complexity is: Space complexity: $O(E(n+k)l)$(on CPU); Time complexity: $O(tknEl)$.
    \item[$\bullet$] \textbf{Extracting the representation of typical samples in all models}: We use $m$ typical samples to perform forward inference on the $N$ models to be merged, and save the hidden layer information of all layers. We adopt the same strategy as step (1), so the space complexity and time complexity are: Space complexity: $O(b 'ElN)$($b'$ will be smaller than $b$ because a typical sample set generally does not occupy all GPU memory); Time complexity: $O((\alpha l\frac{m}{bs}+\beta mEl\frac{s-1}{s}+\gamma mEl)N)$.
    \item[$\bullet$] \textbf{Similarity calculation}: We compute the similarity between the representations of the main model and all other models at each layer, using \textbf{rbf kernel} as the similarity metric, and averaging the similarity across all typical samples, so: Space complexity: $O(mlN)$; Time Complexity: $O(mlEN)$.
    \item[$\bullet$] Obviously, when $n\gg m$, utilizing typical samples will \textbf{greatly reduce the memory and time consumption of extracting features and calculating similarity} (step (3/4)), which is one of the important reasons for our choice of sampling. 
    \item[$\bullet$] \textbf{Hierarchical merging}: Compared with other model merging algorithms, we increase the number of weights in model merging (each model has an independent floating-point weight in each layer), but still take a single parameter as the unit for merging. So: Space complexity: $O(lN)$; Time complexity: $O(N)$.
    \item[$\bullet$] \textbf{Scalability}: When the scale of the model increases, time and space consumption will grow at a \textbf{linear} rate.
\end{itemize}

%% file: tables/sequence_test.tex
% \usepackage{multirow}
\begin{table*}[ht!]
\resizebox{\linewidth}{!}{
\begin{tabular}{c|cc|ccccc|c}
\hline
\centering
\multirow{2}{*}{Method}      & \multicolumn{2}{c|}{\multirow{2}{*}{Task Sequence}} & \multicolumn{5}{c|}{Datasets}                 & \multirow{2}{*}{Average} \\ \cline{4-8}
                             & \multicolumn{2}{c|}{}                               & SST-2  & SQuAD2.0 & MedMCQA & IWSLT2017 & RACE  &                          \\ \hline
\multirow{5}{*}{LoRA SFT}    & Task 1                  & SST-2                      & 95.76 & 31.68   & 32.32   & 13.28     & 44.71 & 43.55                    \\
                             & Task 2                  & SQuAD2.0                   & 94.38 & 87.42   & 16.88   & 25.06     & 58.15 & 56.378                   \\
                             & Task 3                  & MedMCQA                   & 88.3  & 74.89   & 42.62   & 19.29     & 68.38 & 58.696                   \\
                             & Task 4                  & IWSLT2017                 & 76.38 & 75.71   & 42.39   & 45.29     & 58.73 & 59.7                     \\
                             & Task 5                  & RACE                      & 14.79 & 68.05   & 39.8    & 34.85     & 86.24 & 48.746                   \\ \hline
\multirow{5}{*}{EWC}         & Task 1                  & SST-2                      & 95.76 & 31.68   & 32.32   & 13.28     & 44.71 & 43.55                    \\
                             & Task 2                  & SQuAD2.0                   & 94.27 & 88.32   & 25.77   & 20.94     & 51.64 & 56.188                   \\
                             & Task 3                  & MedMCQA                   & 90.47 & 72.12   & 42.53   & 12.44     & 57.75 & 55.062                   \\
                             & Task 4                  & IWSLT2017                 & 81.59 & 65.31   & 41.05   & 47.86     & 55.6  & 58.282                   \\
                             & Task 5                  & RACE                      & 67.42 & 64.81   & 39.68   & 33.54     & 87.34 & \uline{58.558}                   \\ \hline
\multirow{5}{*}{RECALL(Our)} & Task 1                  & SST-2                      & 94.5  & 30.72   & 34.9    & 12.08     & 47.44 & 43.928                   \\
                             & Task 2                  & SQuAD2.0                   & 96.61 & 86.34   & 30.79   & 19.83     & 57.52 & 58.218                   \\
                             & Task 3                  & MedMCQA                   & 92.89 & 71.66   & 40.65   & 18.48     & 69.06 & 58.548                   \\
                             & Task 4                  & IWSLT2017                 & 86.31 & 67.09   & 38.16   & 45.73     & 67.55 & 60.968                   \\
                             & Task 5                  & RACE                      & 80.59 & 62.38   & 36.22   & 43.14     & 88.97 & \textbf{62.26}                    \\ \hline
\end{tabular}
}
\caption{Detailed performance of sequence training scenario.}
\label{Tab:sequence_test}
\end{table*}